\documentclass[sigconf]{acmart}

\AtBeginDocument{%
  \providecommand\BibTeX{{%
    \normalfont B\kern-0.5em{\scshape i\kern-0.25em b}\kern-0.8em\TeX}}}

\usepackage{multirow}
\usepackage{soul}
\usepackage{amsfonts}
\usepackage{lipsum}  
\usepackage{bm}
\usepackage{xcolor}
\usepackage[normalem]{ulem}
\usepackage{xifthen}
\usepackage{pifont}
\usepackage{makecell}
\usepackage{tabularx}
\usepackage{footnote}
\usepackage{subfig}
\makesavenoteenv{tabular}
\makesavenoteenv{table}
\useunder{\uline}{\ul}{}
\usepackage{makecell}
\usepackage{bbm}
\newcommand{\longname}{\textit{\uline{P}ers\uline{o}nalized \uline{M}edical Disease \uline{P}rediction}}
\newcommand{\shortname}{\textit{PoMP}}

\copyrightyear{2024}
\acmYear{2024}
\setcopyright{rightsretained}
\acmConference[WWW '24 Companion]{Companion Proceedings of the ACM Web Conference 2024}{May 13--17, 2024}{Singapore, Singapore}
\acmBooktitle{Companion Proceedings of the ACM Web Conference 2024 (WWW '24 Companion), May 13--17, 2024, Singapore, Singapore}\acmDOI{10.1145/3589335.3651498}
\acmISBN{979-8-4007-0172-6/24/05}

\makeatletter
\gdef\@copyrightpermission{
  \begin{minipage}{0.3\columnwidth}
   \href{https://creativecommons.org/licenses/by/4.0/}{\includegraphics[width=0.90\textwidth]{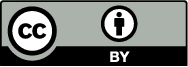}}
  \end{minipage}\hfill
  \begin{minipage}{0.7\columnwidth}
   \href{https://creativecommons.org/licenses/by/4.0/}{This work is licensed under a Creative Commons Attribution International 4.0 License.}
  \end{minipage}
  \vspace{5pt}
}
\makeatother

\begin{document}

\title{Enabling Patient-side Disease Prediction via the Integration of Patient Narratives}

\author{Zhixiang Su}
\affiliation{%
  \institution{Nanyang Technological University}
  \country{Singapore}}
\email{zhixiang002@ntu.edu.sg}

\author{Yinan Zhang}
\affiliation{%
  \institution{Nanyang Technological University}
  \country{Singapore}}
\email{yinan.zhang@ntu.edu.sg}

\author{Jiazheng Jing}
\affiliation{%
  \institution{Nanyang Technological University}
  \country{Singapore}}
\email{jiazheng001@ntu.edu.sg}

\author{Jie Xiao}
\affiliation{%
  \institution{Qilu Hospital of Shandong University}
  \country{China}}
\email{chrisy-4619.202@163.com}

\author{Zhiqi Shen}
\authornote{Corresponding Author}
\affiliation{%
  \institution{Nanyang Technological University}
  \country{Singapore}}
\email{zqshen@ntu.edu.sg}

\renewcommand{\shortauthors}{Zhixiang Su, Yinan Zhang, Jiazheng Jing, Jie Xiao, \& Zhiqi Shen}

\begin{abstract}
  Disease prediction holds considerable significance in modern healthcare, because of its crucial role in facilitating early intervention and implementing effective prevention measures. However, most recent disease prediction approaches heavily rely on laboratory test outcomes (e.g., blood tests and medical imaging from X-rays). Gaining access to such data for precise disease prediction is often a complex task from the standpoint of a patient and is always only available post-patient consultation. To make disease prediction available from patient-side, we propose \longname{} (\shortname{}), which predicts diseases using patient health narratives including textual descriptions and demographic information. By applying \shortname{}, patients can gain a clearer comprehension of their conditions, empowering them to directly seek appropriate medical specialists and thereby reducing the time spent navigating healthcare communication to locate suitable doctors. We conducted extensive experiments using real-world data from \textit{Haodf} to showcase the effectiveness of \shortname{}.
\end{abstract}





\maketitle
\section{Introduction}
Disease prediction has become a highly prioritized and essential aspect in healthcare and related fields in recent years~\cite{Yukun2023}. The ability to forecast illnesses offers invaluable benefits such as early detection and intervention, particularly crucial for conditions like cancer or heart disease where timely treatment is pivotal.  Moreover, predicting chronic diseases (e.g., diabetes) can lead to lifestyle adjustments and timely medications, which potentially halt or mitigate disease progression. Additionally, disease prediction provides invaluable insights into potential health issues before patients seek medical attention, which is particularly beneficial in resource-limited situations. It also benefits patients who, due to limited knowledge of their specific conditions, invest significant time in communication to find the most appropriate doctors.

However, to our best of knowledge, current disease prediction techniques, encompassing both traditional statistical methods~\cite{Botlagunta2023} and advanced deep learning approaches~\cite{Yukun2023}, rely heavily on the data obtained through clinical assessments, including laboratory tests (e.g., blood and urine tests) and diagnostic imaging (e.g., X-rays and CT scans). Unfortunately, such doctor-side comprehensive health data typically become available only after patients engage with healthcare professionals. Consequently, patient-side narratives (e.g., individuals experiencing symptoms) lacking professional terminology and accurate descriptions may face significant challenges in accessing appropriate medical guidance. This challenge is further amplified with the growing popularity of online doctor consultations, a trend accelerated by the Covid-19 pandemic.

To address the outlined challenges and elevate the performance of disease prediction approaches, we propose \longname{} (\shortname{}), which predicts diseases according to patient-side narratives including patient-provided textual descriptions and patient demographic information. \shortname{} enables rapid comprehension of potential health conditions for individuals and seamless connections with doctors specializing in relevant medical disciplines. This innovation simplifies the typically complex process of identifying the appropriate medical department for consultation, thereby significantly reducing the time and effort expended by patients in navigating the healthcare system. 
    
In summary, our contributions are as follows:

\noindent \textbf{Dataset Collection:} To assess the efficacy of \shortname{}, we collected records of patient-doctor consultations from Haodf\footnote{\href{https://www.haodf.com/}{https://www.haodf.com/}\label{ftnt:haodaifu_website}}, a leading online doctor consultation platform in China. Existing publicly available datasets for disease prediction usually focus on the various patient indicators during hospitalization but fall short in capturing patient narratives~\cite{MIMIC3,eicu}. In this work, we acquired narratives from the patient's perspective, including textual descriptions, as well as their basic demographic information (such as age and gender). Additionally, we collected the corresponding diagnoses made by the doctors for further analysis and assessment. We believe that this dataset will serve as a valuable resource for future research.

\noindent \textbf{Patient-side Disease Prediction:} To the best of our knowledge, \shortname{} is the first method capable of predicting a patient's diseases exclusively through patient-side narratives, without relying on any diagnostic test outcomes. \shortname{} presents a promising approach and introduces the possibilities in patient-side disease prediction.

\noindent \textbf{Two-tiered Generic Architecture:} Diseases can be categorized into various levels according to different criteria. Take pneumonia as an example, it can be further broken down into subcategories such as pulmonary nodules, lung adenocarcinoma, etc. To leverage the hierarchical nature of disease classification, we introduce a two-tiered classifier architecture. This method first predicts broad categories and then narrows down to specific disease predictions. Our experimental results on the Haodf dataset have shown that this approach achieves state-of-the-art (SOTA) performance in 6 out of 7 evaluation scenarios.


\section{Methodology}
\subsection{Preliminaries}\label{sec_problem_formulation}
Disease prediction is the process of using patient's medical profiles $M$, for predicting a probable disease $y_i \in Y$. Such medical profiles $M = \{T, C, D\}$ typically contains the following three types of information:
\romannumeral1) Textual descriptions $T$,
\romannumeral2) Numerical continuous data $C$, and
\romannumeral3) Categorical discrete data $D$.
More specifically, we gathered narratives from patients covering various perspectives:


\romannumeral1) Patient-provided textual descriptions $T$: This encompasses text description in natural language that can be obtained from patient self-introductions. It includes chronic disease $t_{\textit{chronic}}$, surgery history $t_{\textit{surgery}}$, radiotherapy history $t_{\textit{therapy}}$, medication usage $t_{\textit{usage}}$, observed symptoms $t_{\textit{symptom}}$, and allergy history $t_{\textit{allergy}}$.

\romannumeral2) Patient demographic information $C$ and $D$: This encompasses basic demographic details including gender $d_{\textit{gender}}$, age $c_{\textit{age}}$, height $c_{\textit{height}}$, weight $c_{\textit{weight}}$, pregnancy situation $d_{\textit{pregancy}}$, and disease duration $c_{\textit{duration}}$.

\subsection{Model Details}

In this work, we propose a generic model, named \longname~(\shortname), to predict diseases according to patient health narratives. We first construct distinct encoders customized for each narrative type. Subsequently, we establish a two-tiered classifier for disease predictions, wherein we first predict the disease category and subsequently the specific disease. Lastly, we discuss about our training regime tailored for the two-tiered generic disease prediction framework.

\subsubsection{Textual Description Encoder}\label{sec_textual_description_encoder}

To effectively capture the semantic knowledge and contextual information in patient-provided textual descriptions, we adopt a Sentence Transformer~\cite{SentenceTransformer} for encoding $T$. Sentence Transformers are pre-trained language models on extensive natural language datasets, capable of considering entire sentences and producing embeddings that encapsulate the overall meaning of the text.

Specifically, we begin by adopting a prompt~\cite{prompt} to better leverage the knowledge learned in a pre-trained language model as follows:
    \begin{equation*}
       \textit{Pro}_{\textit{[TYPE]}} = \textit{[TYPE] is [TEXT]},
    \end{equation*}
where $\textit{[TYPE]}$ is a type of textual descriptions and $\textit{[TEXT]}$ denotes the corresponding textual descriptions $t_{\textit{[TYPE]}} \in T$.

Then, we concatenate all prompt into a unified sentence as follows:
    \begin{equation}
        s= \textit{Concat}(\{\textit{Pro}_{\textit{therapy}}^p,\textit{Pro}_{\textit{usage}}^p,\textit{Pro}_{\textit{surgery}}^p,\textit{Pro}_{\textit{symptom}}^p,\textit{Pro}_{\textit{allergy}}^p\}).
    \end{equation}


Sentence Transformer firstly applies a tokenizer to convert $s$ into tokens $\textit{Token}_t$ and generates an attention mask $\textit{Mask}_t$. Then, Sentence Transformer apply an encoder to convert $\textit{Token}_t$ in to embeddings $\textit{Emb}_t$ as follows:
    \begin{equation}
        \textit{Token}_t, \textit{Mask}_t = \textit{Tokenizer}(s),
    \end{equation}
    \begin{equation}
        \textit{Emb}_t=\textit{Encoder}(\textit{Token}_t).
    \end{equation}

Next, we apply a mean-pooling to reduce the spatial dimensions of feature maps while retaining important information as follows:
    \begin{equation}
        \textit{Emb}_t^{'}= \frac{\textit{Emb}_t * \textit{Mask}_t}{\textit{sum}(\textit{max}(\textit{Mask}_t, \epsilon))},
    \end{equation}
where $\epsilon$ denotes a minimum value to avoid divided by zero.

Lastly, we apply a normalize layer to generate the ultimate textual description embeddings as follows:
    \begin{equation}
        \textit{Emb}_{\textit{text}}= \frac{\textit{Emb}_t^{'}}{\textit{max}(||\textit{Emb}_t^{'}||_2, \epsilon)}.
    \end{equation}

\subsubsection{Demographic Information Encoder}\label{sec_continuous_discrete}

As mentioned in Section~\ref{sec_problem_formulation}, Demographic information is composed of continuous data and discrete data. We handle them through different processes.

For continuous data, we employ normalization to scale the values within the range $[0,1]$, ensuring efficient convergence of model parameters during training as follows:    
    \begin{equation}
        \textit{Emb}_c^{\textit{norm}}= \{\frac{\textit{c}}{\textit{max}(||\textit{c}||_2, \epsilon)}, \textit{c} \in C \}.
    \end{equation}
In the context of patient-side disease prediction, $C$ comprises the following components:
    \begin{equation}
        C^p=\{ c_{\textit{age}}, c_{\textit{height}}, c_{\textit{weight}}, c_{\textit{duration}}\}.
    \end{equation}

For discrete data, we apply one-hot embeddings as follows:
    \begin{equation}
        \textit{Emb}_d^{\textit{norm}} = \{\textit{Embedding}(\textit{OneHot}(d)),\textit{d} \in D \}.
    \end{equation}
For patient-side disease prediction, $D=\{d_{\textit{gender}}, d_{\textit{pregancy}}\}$.


Subsequently, both continuous and discrete data undergo encoding via a multi-head attention layer as follows:
    \begin{equation}
        Q, K, V = \textit{Linear}_{Q,K,V} (\textit{Concat}(\textit{Emb}_c^{\textit{norm}} \cup \textit{Emb}_d^{\textit{norm}})).
    \end{equation}
    \begin{equation}
        \textit{Emb}_{\textit{data}}=\textit{Concat}(\{\textit{head}_1,...,\textit{head}_n\})W^o,
    \end{equation}
where $\textit{head}_i = \textit{Attention}(QW_i^Q,KW_i^K,VW_i^V)$.

\subsubsection{Patient-side Disease Prediction Classifier}\label{sec_classifier}

Given the hierarchical structure of diseases categorization, we propose a two-tier classification approach. First, we categorize diseases into broader categories: 
    \begin{equation}
        \textit{Emb}_{\textit{all}}=\textit{Normalize}(\textit{Linear}(\textit{Concat}(\{{\textit{Emb}_{\textit{data}}} \cup \textit{Emb}_{\textit{text}}\})))),
    \end{equation}
    \begin{equation}
        y'_{\textit{cate}}=\textit{Softmax}(\textit{Emb}_{\textit{all}}).
    \end{equation}

After predicting the category, the model can narrow down to the potential diseases, thereby simplifying the subsequent prediction task. The category with the highest score, denoted as $y'^{\textit{max}}_{\textit{cate}}$, is selected as the candidate category. Subsequently, we apply a category-specific $\textit{Softmax}()$ to predict the specific disease within the chosen category:

    \begin{equation}
        \textit{Emb}_{y'^{\textit{max}}_{\textit{cate}}}=\textit{Normalize}(\textit{Linear}_{y'^{\textit{max}}_{\textit{cate}}}(\textit{Concat}(\{{\textit{Emb}_{\textit{data}}} \cup \textit{Emb}_{\textit{text}}\})))),
    \end{equation}
    \begin{equation}
        y'_{\textit{dise}}=\textit{Softmax}_{y'^{\textit{max}}_{\textit{cate}}}(\textit{Emb}_{\textit{cate}}),
    \end{equation}
where $\textit{Linear}_{y'^{\textit{max}}_{\textit{cate}}}$ and $\textit{Softmax}_{y'^{\textit{max}}_{\textit{cate}}}$ are category-specific linear and softmax functions, respectively.

\subsubsection{Training}\label{sec_training}

After receiving the category prediction $y'_{\textit{cate}}$ and disease prediction $y'_{\textit{dise}}$, we define the training objective and loss function as follows:

\noindent \textbf{Objective:} Because different categories may contain overlapping disease, incorrect category predictions can still lead to correct disease predictions. However, for the prediction chain to align with human cognition, we only consider predictions as correct if both the category and disease are accurately predicted.

\noindent \textbf{Loss function:} To integrate category prediction loss with disease prediction loss, we utilize a weighted cross-entropy loss defined as follows:
\begin{equation}
    \textit{CrossEntropy}(y,y')= \sum_{i=1}^M y_i* \textit{log}(y'_i),
\end{equation}

\begin{equation}
    \textit{Loss}= \textit{CrossEntropy}(y_{\textit{cate}},y'_{\textit{cate}})+\alpha*\textit{CrossEntropy}(y_{\textit{dise}},y'_{\textit{dise}}),
\end{equation}
where $M$ denotes the number of category (or disease) labels and $\alpha$ is a weight hyper-parameter.

\begin{table}[!t]
    \centering
    \caption{Statistics of the Haodf dataset.}
    \label{tab:dataset_Statistic}
    \resizebox{\columnwidth}{!}{
   \begin{tabular}{ccccccc|c} \toprule
    Category    & \textit{Cold}    & \textit{Diab.}& \textit{CHD}& \textit{Depr.}    & \textit{Pneu.}    & \textit{Lung.}    & All   \\ \midrule
    \# Records       & 1413 & 4157 & 4543 & 4965 & 6143 & 9518 & 29326 \\
    \# Disease & 29   & 41   & 63   & 31   & 29   & 55   & 190  \\ 
    \# Avg. Token per Patient &  403.9  &  389.9  & 575.3   &  165.4  &  548.7  & 595.2   &  481.4 \\ 
    \bottomrule
\end{tabular}
    }
\end{table}

\begin{table*}[!t]
\centering
\caption{Category prediction and disease prediction results on Haodf dataset.} \label{table_main_results}
\vspace{-0.2cm}
\begin{tabular}{llccccccc}
\toprule
        &        & gpt2  & bert-base & t5-small & albert-base-v2 & electra-small & roberta-base   & PoMP           \\         \midrule

\multirow{3}{*}{Category} & Hit@1  & 0.695                    & 0.806                         & 0.798                        & 0.806                              & 0.799                             & 0.802                            & \textbf{0.811}           \\
         & Hit@3  & 0.939                    & 0.968                         & 0.973                        & 0.968                              & 0.965                             & 0.974                            & \textbf{0.979}           \\
         & AUC-PR & 0.797                    & 0.837                         & 0.837                        & 0.832                              & 0.819                             & \textbf{0.838}                   & 0.836                \\
         \midrule \midrule
\multirow{3}{*}{Disease}  & Hit@1  & 0.104                    & 0.102                         & 0.111                        & 0.115                              & 0.085                             & 0.097                            & \textbf{0.135}           \\
         & Hit@3  & 0.107                    & 0.111                         & 0.128                        & 0.124                              & 0.087                             & 0.101                            & \textbf{0.151}           \\
         & Hit@10 & 0.115                    & 0.142                         & 0.139                        & 0.128                              & 0.095                             & 0.115                            & \textbf{0.167}           \\
         & AUC-PR & 0.089                    & 0.105                         & 0.103                        & 0.101                              & 0.077                             & 0.094                            & \textbf{0.119}    \\ \bottomrule
\end{tabular}
\end{table*}

\begin{table}[!t]
\centering
\caption{Ablation study results compared to vanilla Sentence Transformer.}\label{table_ablation_study}

\begin{tabular}{lcccc}\toprule
\multicolumn{1}{c}{} & \multicolumn{1}{l}{Hit@1} & \multicolumn{1}{l}{Hit@3} & \multicolumn{1}{l}{Hit@10} & \multicolumn{1}{l}{AUC-PR} \\
\midrule
\multicolumn{5}{c}{\textbf{Category}}                                                                                                           \\
\midrule
Text Only            & 0.804                     & 0.983                     & 1.000                      & 0.830                      \\
PoMP (Ours)          & 0.811                     & 0.979                     & 1.000                      & 0.836                      \\
\midrule
\multicolumn{5}{c}{\textbf{Disease}}                                                                                                            \\
\midrule
Text Only            & 0.111                     & 0.118                     & 0.125                      & 0.111                      \\
PoMP (Ours)          & 0.135                     & 0.151                     & 0.167                      & 0.119              \\ \bottomrule       
\end{tabular}
\end{table}



\section{Dataset}
To evaluate the effectiveness of \shortname{}, we created the \textit{Haodf} dataset. We collected comprehensive patient-doctor consultation records including patient-side narratives across six prevalent disease categories, which were further classified based on their associated risk levels. These categories (see Table \ref{tab:dataset_Statistic}) include \romannumeral1) low-risk categories: \textit{Common Cold} (\textit{Cold}) and \textit{Pneumonia} (\textit{Pneu.}); \romannumeral2) medium-risk categories: \textit{Diabetes} (\textit{Diab.}) and \textit{Depression} (\textit{Depr.}); and \romannumeral3) high-risk categories: \textit{Coronary Heart Disease} (\textit{CHD}) and \textit{Lung Cancer} (\textit{Lung.}).

\begin{figure}
    \centering
    \subfloat[Patient gender distribution.]{\includegraphics[width=0.48\columnwidth]{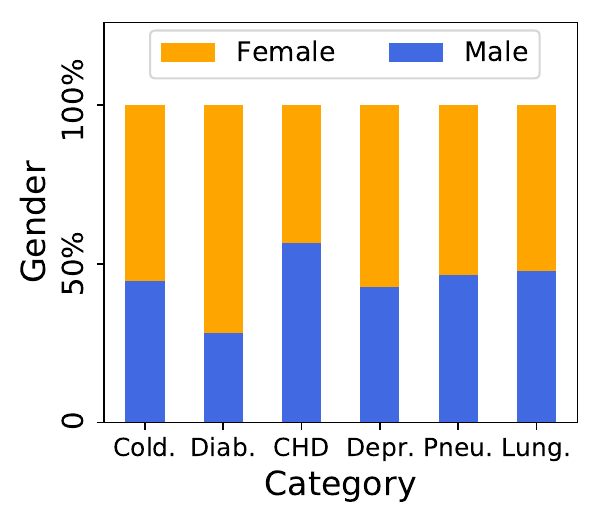}\label{fig:fig_sta_gender_for_all_disease}} \hfill
    \subfloat[Patient age distribution.]{\includegraphics[width=0.48\columnwidth]{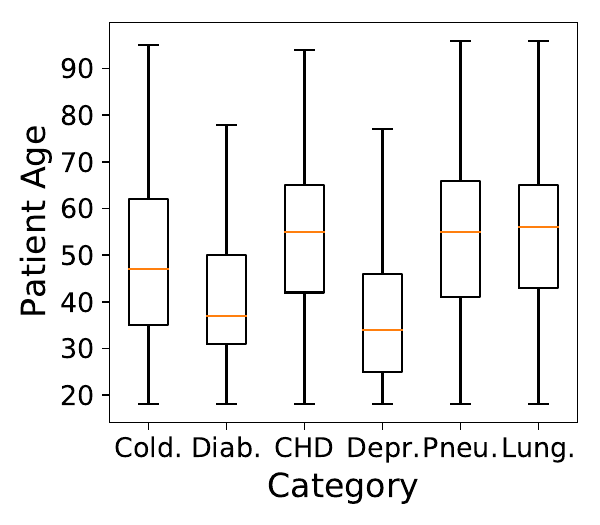}\label{fig:sta_age_for_all_disease}}
    \caption{Distribution of patient demographics across six categories.}
    \label{fig:dataset_sta}
\end{figure}

To demonstrate the potential correlation between disease and patient demographics, we conducted an analysis of gender and age distributions across all six disease categories (see Figure \ref{fig:fig_sta_gender_for_all_disease} and Figure \ref{fig:sta_age_for_all_disease}). We observed distinct variations in susceptible populations across different diseases.

\section{Experiment}
\subsection{Baselines}
In our experiments, we compare \shortname{} against six widely-adopted Natural Language Processing (NLP) models. These models are standard implementations of pre-trained language models (PLMs) including GPT2~\cite{GPT2}, BERT~\cite{BERT}, T5~\cite{T5}, ALBERT~\cite{ALBERT}, ELECTRA~\cite{ELECTRA}, and RoBERTa~\cite{RoBERTa}.  We apply the two-tiered classification approach outlined in Section~\ref{sec_classifier} to make disease predictions with \shortname{}. For all baseline models, category predictions and disease predictions are performed independently.

We implement \shortname{} based on the SOTA Sentence Transformer (all-MiniLM-L6-v2\footnote{huggingface.co/sentence-transformers/all-MiniLM-L6-v2}). 
The training process utilized two NVIDIA Tesla V100 GPUs equipped with 32GB RAM.
We have made both the dataset and the source code publicly available\footnote{https://github.com/ZhixiangSu/PoMP}.

\subsection{Results and Analysis}
The results of category predictions and disease predictions are presented in Table~\ref{table_main_results}. We utilize the hit rate (Hit@k) and the area under the precision-recall curve (AUC-PR) for evaluation, with the best performance highlighted in bold.

In category predictions, \shortname{} achieves the highest performance in terms of Hit@1 and Hit@3, while its AUC-PR scores are comparable to PLM baselines. The notably strong performance in category prediction further supports the two-tiered classification strategy we proposed.


In disease predictions, \shortname{} achieves the highest performance in terms of all metrics among all baselines. Notably, the substantial improvement relative to the second-best approaches approaches are $+17.3\%$ for Hit@1, $+18.0\%$ for Hit@3, $+17.6\%$ for Hit@10, and $+13.3\%$ for AUC-PR. These findings highlight the efficacy of our two-tiered classification strategy.

\subsection{Ablation Study}

To demonstrate the significance of demographic information in disease prediction, we conducted an ablation study. This study compare \shortname{} to the vanilla Sentence Transformer model, which accepts text-only inputs.


The result of category predictions and disease predictions are illustrated in Table~\ref{table_ablation_study}. \shortname{} achieves a better results in terms of Hit@1 and AUC-PR for category prediction. For disease prediction, \shortname{} achieves a significant better result in terms of all metrics. Notably, the vanilla Sentence Transformer appears to suffer from limited discriminatory capacity, as indicated by the small performance gaps among Hit@1, Hit@3, and Hit@10 (increases of $+7.2\%$ and $+5.6\%$). In contrast, \shortname{} exhibits larger performance disparities, with improvements of $+11.8\%$ and $+9.5\%$, respectively.

\section{Conclusion}
In conclusion, we address the critical need for early disease prediction by introducing \longname{} (\shortname{}), an innovative approach that leverages only patient-provided health narratives through a two-tiered prediction model. \shortname{} simplifies the process of connecting patients with appropriate medical specialists, representing a substantial advancement in making disease prediction more accessible and tailored to patient needs, thereby enhancing the efficiency of healthcare communication. To validate the effectiveness of \shortname{}, we collected extensive patient-doctor consultation records from the Haodf platform, encompassing a wide array of patient narratives detailing their conditions. We believe this work will lay a solid groundwork for future research in patient-side disease prediction.

\section{Acknowledgment}
This research is supported, in part, by A*STAR under its RIE2025 Industry Alignment Fund – Industry Collaboration Projects (IAF-ICP) Funding Initiative (Award No: I2301E0026); the Joint NTU-UBC Research Centre of Excellence in Active Living for the Elderly (LILY); Alibaba-NTU Singapore Joint Research Institute (JRI), Nanyang Technological University, Singapore. This research is also supported, in part, by the National Research Foundation, Prime Minister’s Office, Singapore under its NRF Investigatorship Programme (NRFI Award No. NRF-NRFI05-2019-0002). Any opinions, findings and conclusions or recommendations expressed in this material are those of the authors and do not reflect the views of National Research Foundation, Singapore.

\bibliographystyle{ACM-Reference-Format}
\bibliography{sample-base}
\end{document}